\documentclass[11pt]{article}

\usepackage[utf8]{inputenc}
\usepackage{amsmath, amssymb}
\usepackage{graphicx}
\usepackage{hyperref}
\usepackage{geometry}
\usepackage{authblk}
\usepackage{tabularx}

\title{Graph Neural Networks for Heart Failure Prediction on an EHR-Based Patient Similarity Graph}

\author[1,2]{Heloisa Oss Boll}
\author[2]{Ali Amirahmadi}
\author[2]{Amira Soliman}
\author[2]{Stefan Byttner}
\author[1,3]{Mariana Recamonde-Mendoza}

\affil[1]{Institute of Informatics, Universidade Federal do Rio Grande do Sul (UFRGS), Porto Alegre, Brazil}
\affil[2]{School of Information Technology, Halmstad University, Halmstad, Sweden}
\affil[3]{Bioinformatics Core, Hospital de Clínicas de Porto Alegre (HCPA), Porto Alegre, Brazil}

\date{}

\begin{document}

\maketitle

\begin{abstract}
\textbf{Objective:} In modern healthcare, accurately predicting diseases is a crucial matter. This study introduces a novel approach using graph neural networks (GNNs) and a Graph Transformer (GT) to predict the incidence of heart failure (HF) on a patient similarity graph at the next hospital visit. 

\textbf{Materials and Methods:} We used electronic health records (EHR) from the MIMIC-III dataset and applied the K-Nearest Neighbors (KNN) algorithm to create a patient similarity graph using embeddings from diagnoses, procedures, and medications. Three models — GraphSAGE, Graph Attention Network (GAT), and Graph Transformer (GT) — were implemented to predict HF incidence. Model performance was evaluated using F1 score, AUROC, and AUPRC metrics, and results were compared against baseline algorithms. An interpretability analysis was performed to understand the model's decision-making process.

\textbf{Results:} The GT model demonstrated the best performance (F1 score: 0.5361, AUROC: 0.7925, AUPRC: 0.5168). Although the Random Forest (RF) baseline achieved a similar AUPRC value, the GT model offered enhanced interpretability due to the use of patient relationships in the graph structure. A joint analysis of attention weights, graph connectivity, and clinical features provided insight into model predictions across different classification groups.

\textbf{Discussion and Conclusion:} Graph-based approaches such as GNNs provide an effective framework for predicting HF. By leveraging a patient similarity graph, GNNs can capture complex relationships in EHR data, potentially improving prediction accuracy and clinical interpretability.
\end{abstract}


\vspace{9em}
\noindent \textbf{*Corresponding authors:} \\
Email: \texttt{hoboll@inf.ufrgs.br}, Phone: +55 51 3308-6843; \\
Email: \texttt{mrmendoza@inf.ufrgs.br}, Phone: +55 51 3308-6843


\section{Introduction}
With the advancement of electronic health record (EHR) systems and artificial intelligence techniques, the healthcare sector has undergone a significant positive transformation \cite{bohrRiseArtificialIntelligence2020}. Clinical risk prediction models are increasingly used to identify individualized risks, including complex diseases such as heart failure (HF)  \cite{sharmaClinicalRiskPrediction2021a, yinDomainKnowledgeGuided2019}. 

Deep learning models contribute to an essential part of this success. They can independently identify and extract important features from data, allowing the processing of dense datasets such as EHRs and thus leading to improved predictive performance \cite{xiaoOpportunitiesChallengesDeveloping2018}. For example, DoctorAI and Dipole utilized recurrent neural networks (RNNs) and historical EHR patient data to make predictions \cite{choiDoctorAIPredicting2016, maDipoleDiagnosisPrediction2017}, while Deepr used a convolutional neural network (CNN) for EHR-based risk stratification \cite{nguyenDeeprConvolutionalNet2016}. 

One major pitfall is that most of these architectures disregard the relational information underlying EHR data, treating medical information as a ﬂat-structured bag of features \cite{choiLearningGraphicalStructure2020}. Graph-based deep learning methods, such as graph neural networks (GNNs), have been employed in healthcare and other areas to make predictions on data from non-Euclidean domains \cite{bronsteinGeometricDeepLearning2017}.  

This can be achieved by modeling patient data as graphs, where nodes represent clinical entities like patients, diagnoses, or treatments, and edges capture relationships such as co-occurring conditions or shared treatments. For instance, the HarmOnized Representation learning on Dynamic EHR graphs (HORDE) model utilized a multimodal dynamic EHR graph for multiple patient-related tasks, and MedPath aimed to extract personalized subgraphs from medical ontology graphs for individual risk predictions \cite{yeMedPathAugmentingHealth2021}.

Although studies on GNNs and EHR graphs have been performed, this area remains largely unexplored, especially considering patient similarity graphs, with few observed works \cite{rocheteauPredictingPatientOutcomes2021, golmaeiDeepNoteGNNPredictingHospital2021, pieroniPredictiveModelingApplied2021, anMERGEMultigraphAttentive2022, zhangPM2F2NPatientMultiview, tangPredicting30dayAllcause2023, tariq2023graph}. This article aims to extend current research by predicting HF using a patient similarity graph constructed from medication, diagnosis, and procedure codes. Furthermore, we introduce a graph-based interpretability analysis to understand prediction patterns, helping clinicians make better use of the learned similarity insights.

Specifically, our contributions to the field include: (1) developing a novel methodology for constructing patient similarity graphs using dense, pretrained representations of multivariate EHR data; (2) conducting an evaluation and benchmarking of three GNN architectures — GraphSAGE, GAT, and GT — for HF diagnosis, addressing the limited GNN model comparisons in existing works \cite{tariq2023graph, tangPredicting30dayAllcause2023, pieroniPredictiveModelingApplied2021}; (3) performing a detailed ablation study to investigate the relevance of clinical features for HF prediction; and (4) introducing an in-depth interpretability framework analyzing graph descriptive statistics, attention weights, and medical features, expanding on the limited graph interpretability seen in previous studies \cite{zhangPM2F2NPatientMultiview, golmaeiDeepNoteGNNPredictingHospital2021, rocheteauPredictingPatientOutcomes2021}.

\section{Materials and Methods}
\subsection{Data sources}
The study was based on the Medical Information Mart for Intensive Care III dataset, or MIMIC-III \cite{johnson_mimic-iii_2015}. Diagnoses and procedures were encoded with the ICD-9 ontology \cite{InternationalClassificationDiseases}, while medications with NDC. Patient data were processed using Pandas and the PyHealth library \cite{pyhealth23}, which organized the EHR information into a structured dictionary format. We included patients who had at least two hospital visits to allow for the prediction of HF on a subsequent visit, resulting in a final sample of 4,760 patients with 8,891 unique visits. The total number of features was 4788, with 817 diagnosis codes, 517 procedure codes, and 3454 prescription codes. Our dataset was imbalanced, with about 28\% of patients with HF (i.e., positive class). 

Labeling for HF was based on the presence of ICD-9 codes for HF during patient visits, following guidelines from the New York State Department of Health’s ICD-9 Workbook \cite{NYHealthHF2024}. If a heart failure code was found, we excluded that visit and all subsequent ones to prevent data leakage and labeled the patient as positive (1). Patients without any recorded HF diagnosis were labeled as negative (0). Further information is available in the Supplementary Materials.

\subsection{Patient representation}
We employed pre-trained low-dimensional medical embeddings to create patient representations. These consisted of 300-dimensional vectors for ICD-9 and NDC medication codes, generated with skip-gram \cite{choiLearningLowDimensionalRepresentations2016}. The method captured relationships between medical codes by predicting the likelihood of their co-occurrence within a large corpus of healthcare claims data. The resource is publicly available on GitHub \cite{ClinicalMLEmbeddings}.

\begin{figure}[h]
    \centering
    \includegraphics[width=0.85\linewidth]{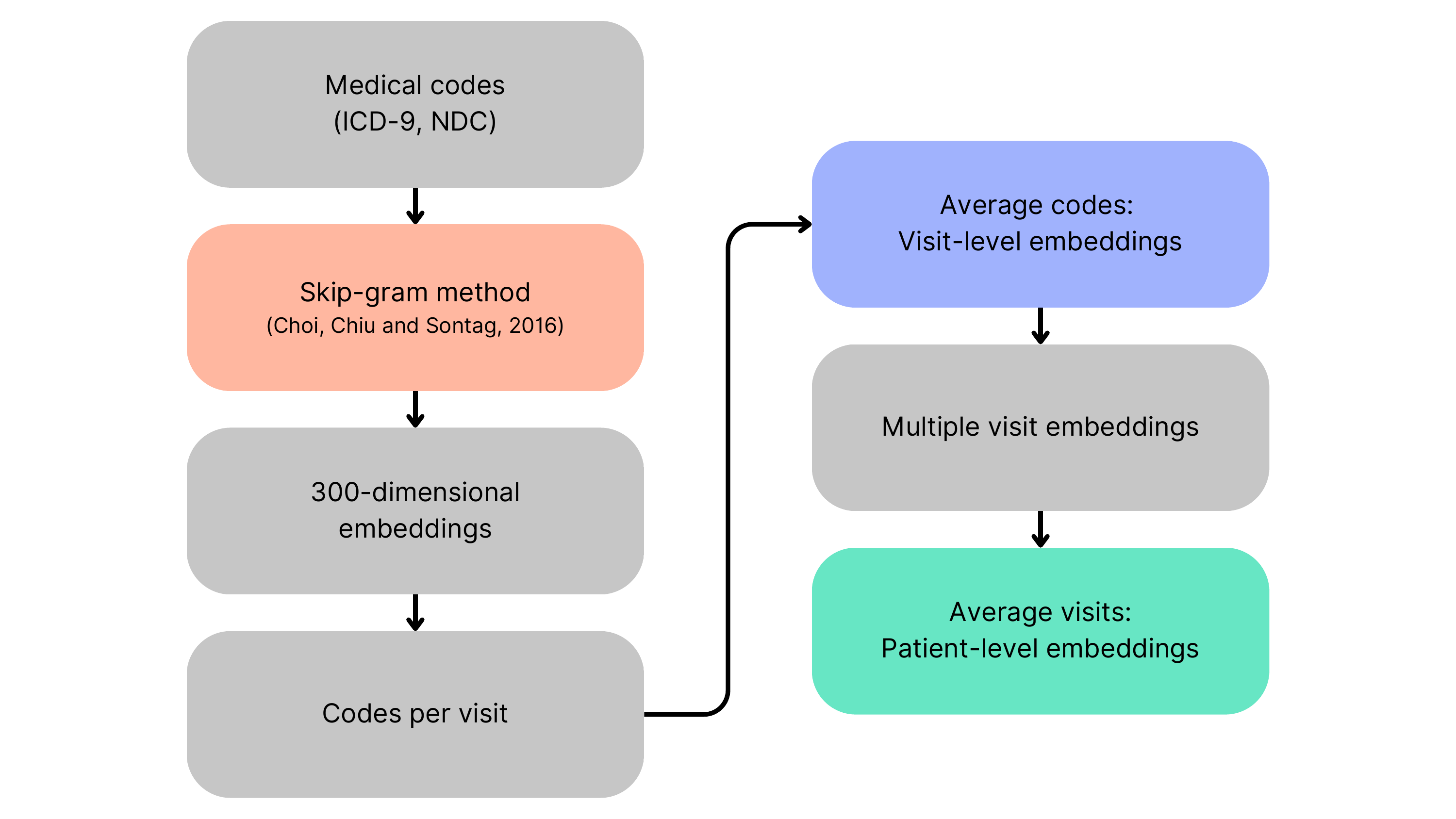} 
    \caption{Process for generating visit-level and patient-level representations based on medical concept embeddings.}
    \label{fig:workflow}
\end{figure}

We first extracted the sets of diagnoses, procedures, and medication codes recorded during the patient's hospital visits. These were mapped to their corresponding embeddings, and an average of these embeddings was computed to represent each visit. We then averaged these visit-level embeddings to form a unique representation for each patient (\autoref{fig:workflow}). These were utilized to both construct the patient similarity graph and as input features in the predictive models.

\subsection{Patient similarity graph}

Similarity between patient feature vectors was quantified using cosine similarity. We applied the K-Nearest Neighbors (KNN) algorithm for a range of K's (2-10) to the similarity matrix to determine the optimal number of edges for each patient node.

K = 3 was chosen based on the distortion metric, which measures the sum of squared distances between each point and its nearest centroid. Consequently, each patient node was linked to its three most similar neighbors in a NetworkX graph \cite{networkx}. We noted, however, that a given node could have more than three neighbors if it was selected as the most similar node by multiple other nodes.

\begin{figure}[h]
    \centering
    \includegraphics[width=0.7\linewidth]{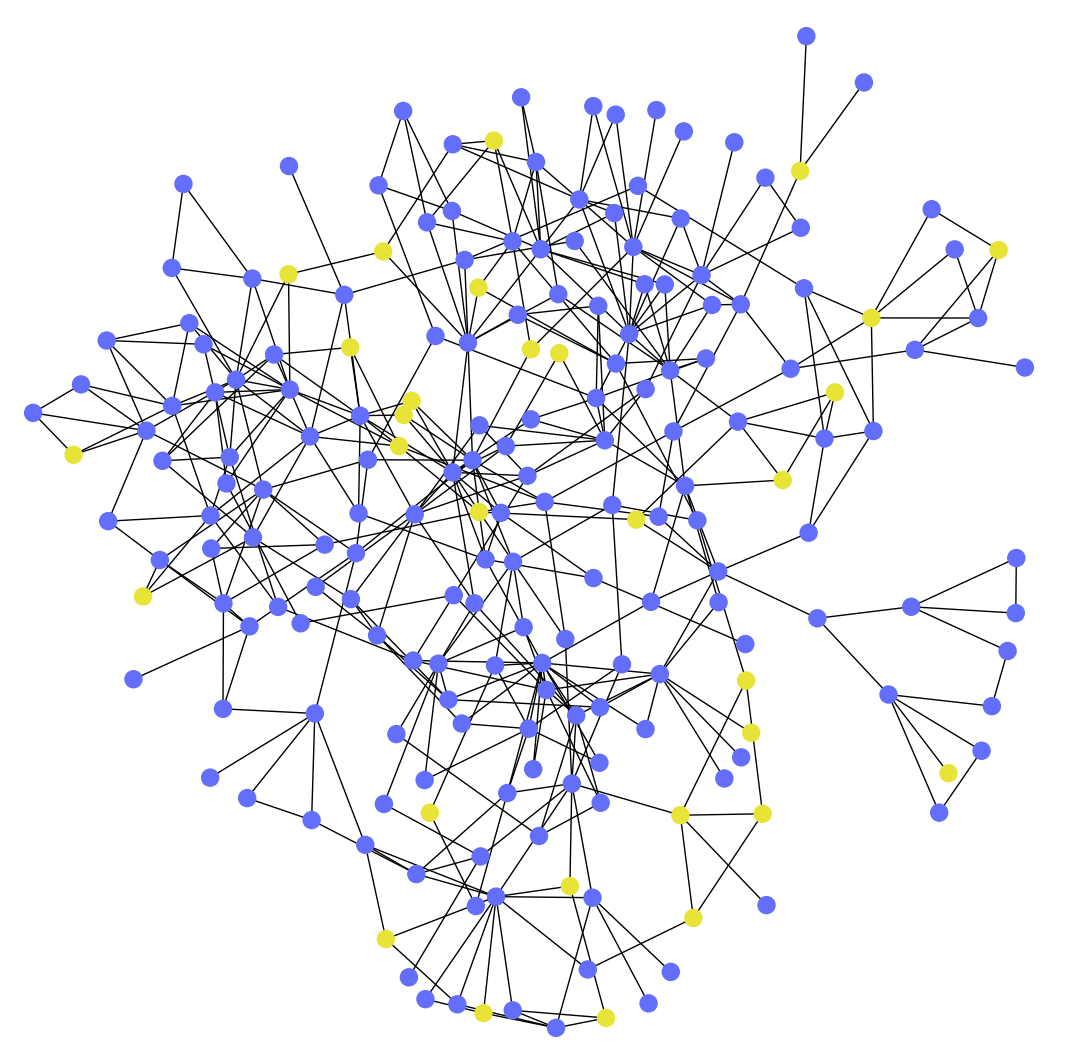} 
    \caption{Visualization of a random subset of nodes from the similarity graph, based on a breadth-first search and the Kamada-Kawai algorithm. Yellow dots represent patients with HF (i.e., positive cases), while blue represents other patients (i.e., negative cases).}
    \label{fig:graph}
\end{figure}

The final graph included all 4,760 patients, with each node carrying the original patient-level embedding as a feature. A subgraph of 200 nodes can be found in  (\autoref{fig:graph}).

The graph was split into training, validation, and test sets (60-20-20) using the DeepSNAP library \cite{deepsnap}. As DeepSNAP only supports fixed splits in graph transductive learning,  we used a fixed data split across experiments. A summary of the graph data and each set is available in \autoref{tab:graphinfo}.

\begin{table}[h]
    \centering
    \caption{Details from the constructed patient similarity graph. Nodes were split while maintaining the class imbalance ratio, ensuring the proportion from the full graph.}
    \resizebox{\linewidth}{!} {
    \begin{tabularx}{\linewidth}{|l|X|X|X|X|} 
        \hline 
        \textbf{Characteristics} & \textbf{Full graph} & \textbf{Train mask} & \textbf{Val. mask} & \textbf{Test mask} \\ 
        \hline 
        \# total nodes & 4760 & 2856 & 952 & 952 \\ 
        \hline 
        \# edges & 11763 & 11763 & 11763 & 11763 \\ 
        \hline 
        \# positive nodes & 1062 & 633 & 215 & 214 \\ 
        \hline 
        \# negative nodes & 3698 & 2223 & 737 & 738 \\ 
        \hline 
        \% positive instances & 28.71 & 28.47 & 29.17 & 28.99 \\ 
        \hline
    \end{tabularx}
    }
    \label{tab:graphinfo}
\end{table}

\subsection{Model architectures and implementation}
We selected GraphSAGE (SAGE) \cite{hamiltonInductiveRepresentationLearning}, which learns node representations by sampling and aggregating features from their local neighborhoods; Graph Attention Network (GAT) \cite{velickovicGraphAttentionNetworks2018}, which introduces attention to weigh the importance of neighboring nodes for a given node's new representation; and Graph Transformer (GT) \cite{Shi2020UniMP}, based on a more advanced attention mechanism. These were implemented with PyTorch Geometric (PyG) \cite{fey2019pyg} and trained to perform binary node classification at the threshold of 0.5. Batch normalization was utilized to stabilize learning. 

All experiments were repeated thrice over the same split and conducted on an Nvidia RTX 6000 GPU. Hyperparameter optimization was performed with Optuna and Weights \& Bias \cite{optuna, wandb}. Early stopping was incorporated to prevent overfitting. All code is available on \href{https://github.com/hossboll/patient-gnn}{https://github.com/hossboll/patient-gnn}.

\subsection{Evaluation}
 The F1 score was selected as the primary metric for both GNN optimization and evaluation, as it balances precision and recall. Models were selected based on the highest F1 scores over the validation set. Other evaluation metrics include the Area Under the Precision-Recall Curve (AUPRC), Area Under the Receiver Operating Characteristic Curve (AUROC), Accuracy (Acc), Balanced Accuracy (Bal. Acc), Precision (Prec; also referred to as positive predictive value (PPV)), and Recall (Rec; also referred to as sensitivity). 

For benchmarking, we compared the performance of the best GNN model, the GT, with hyperparameter-tuned Random Forest (RF), K-Nearest Neighbors (KNN), Logistic Regression (LR), Gradient Boosting Trees (GBT), and a deep Multilayer Perceptron (MLP), implemented using Scikit-learn \cite{scikit-learn}. To ensure consistency, we first identified the patient nodes masked in the training, validation, and test sets within the similarity graph. We then used their corresponding node features as inputs in the baseline models.

\subsection{Interpretability}
To interpret the prediction patterns of the GT model, we examined three axes: graph connectivity patterns, attention weights, and clinical features within the patient similarity graph across the four classification groups — true positives (TP), true negatives (TN), false positives (FP), and false negatives (FN). After, we performed an integrative analysis over four random instances, one from each group.

\section{Quantitative results}
\subsection{GNN architecture performance}
\begin{figure}[h]
    \centering
    \includegraphics[width=1\textwidth]{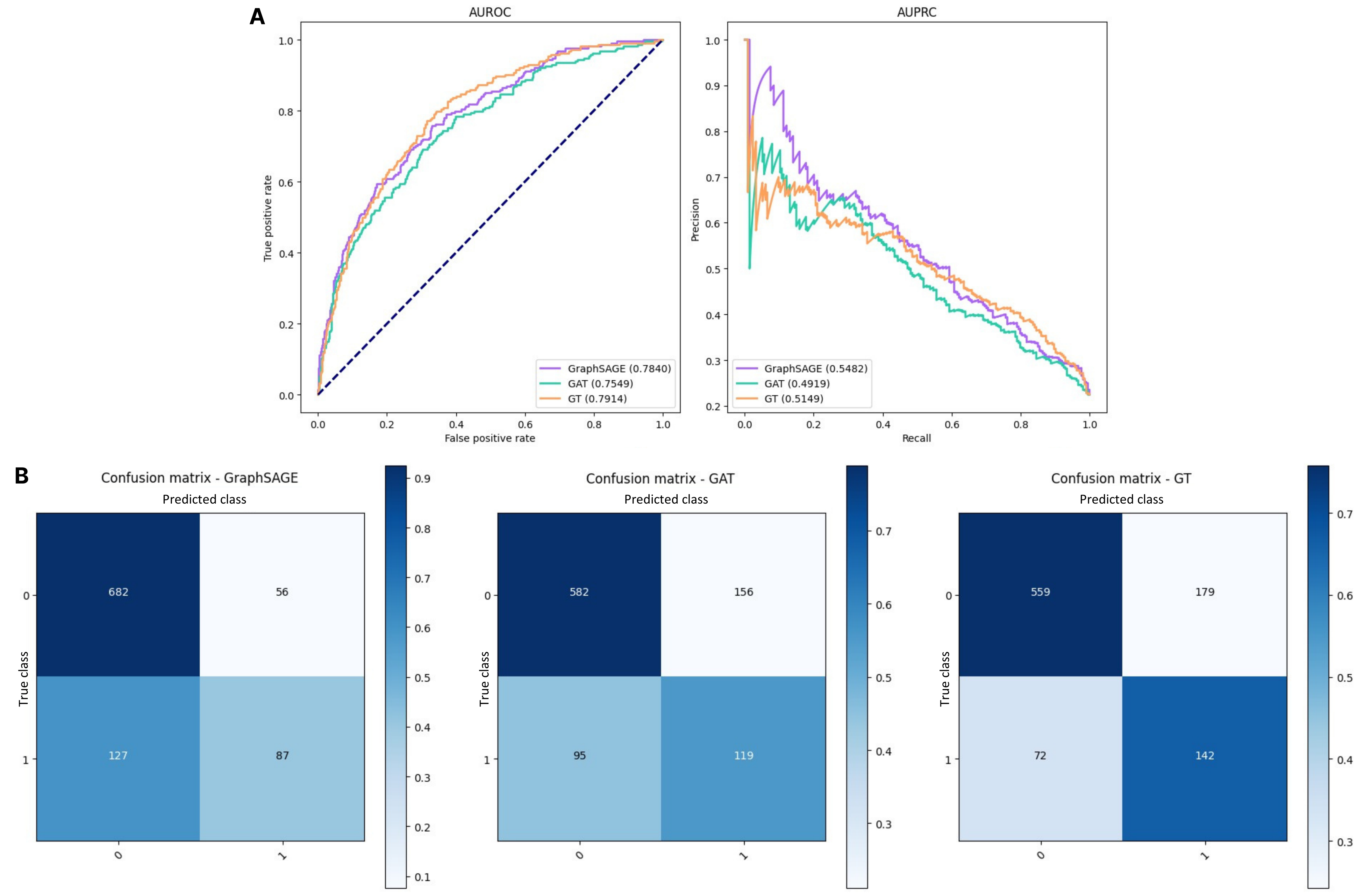}
    \caption{AUROC and AUPRC curves (A) and confusion matrices (B) for GraphSAGE, GAT, and GT models on the test set. GraphSAGE has a higher true negative rate, while GT shows better recall, which is important for the detection of positive HF cases.}
    \label{fig:cms-bce}
\end{figure}

First, we aimed to investigate which GNN architecture performed best in predicting HF. The GT model achieved the highest F1 test score (0.5328), although GraphSAGE showed the highest AUPRC (0.5476)  (\autoref{tab:bce-test}). Confusion matrices and AUROC/AUPRC curves detail the predictions, indicating the GT's improved ability to identify positive cases (\autoref{fig:cms-bce}). 
 
 Secondly, we replaced the binary cross-entropy with other loss functions designed for class-imbalanced scenarios, the weighted binary cross-entropy (WBCE) and the focal (FL) losses. The GT model with FL ($\alpha=0.75$, $\gamma=1$) achieved the best metrics (F1 score: 0.5531, AUROC: 0.7914, AUPRC: 0.5393). Detailed hyperparameters and training and validation data can be found in the Supplementary Materials.

\begin{table}[h]
\caption{Test results from GNN models optimized with the BCE loss, each run thrice. The GT model shows the highest F1 and recall scores. Standard deviations indicate variations across runs considering the same graph split.}
\centering
\resizebox{\linewidth}{!} {
\begin{tabularx}{\linewidth}{|l|X|X|X|X|}
\hline
\textbf{Metric}& \textbf{SAGE}& \textbf{GAT}& \textbf{GT}\\ 
\hline
F1& 0.4758 ± 0.011& 0.4832 ± 0.003& \textbf{0.5328 ± 0.003}\\ 
\hline
Acc              & \textbf{0.8032} ± 0.004& 0.7356 ± 0.000& 0.7377 ± 0.002\\ 
\hline
Bal. Acc     & 0.6591 ± 0.006& 0.6697 ± 0.002& \textbf{0.7112 ± 0.002}\\ 
\hline
Rec& 0.3972 ± 0.008& 0.5498 ± 0.005& \textbf{0.6651 ± 0.002}\\ 
\hline
Prec             & \textbf{0.5931 ± 0.017}& 0.4310 ± 0.001& 0.4443 ±  0.003\\ 
\hline
AUROC                 & 0.7824 ± 0.000& 0.7537 ± 0.001& \textbf{0.7918 ± 0.002}\\ 
\hline
AUPRC                 & \textbf{0.5476 ± 0.00}1& 0.4931 ± 0.001& 0.5200 ± 0.002\\
\hline
\end{tabularx}
}
    \label{tab:bce-test}
\end{table}

\subsection{Impact of clinical data}
To evaluate the impact of each data type on HF prediction, we retrained the GT with FL model using only medication, procedure, or diagnosis data. Medication data alone resulted in the highest recall, followed by diagnosis, with procedures having the least impact. The use of all three data types achieved the best performance. Details are available in the Supplementary Materials.

Next, we conducted an ablation study by removing one data source at a time. The results confirmed that excluding medication data led to the most significant performance drop, followed by diagnosis data. Removing procedure data had the least impact. The combined model achieved the highest F1 score (0.5361) and AUPRC (0.5227), highlighting the importance of integrating multiple data sources. A summary is provided in \autoref{tab:ablation}.

\begin{table*}[h]
\centering
\caption{Test results from the GT models with FL ($\alpha=0.75$, $\gamma=1$) for the ablation study, each run thrice. Using data from the three sources results in a superior predictive performance. Standard deviations indicate variations across runs considering the same graph split.}
\resizebox{\textwidth}{!}{%
\begin{tabular}{|c|c|c|c|c|} 
\hline
\textbf{Metric (Test)} & \textbf{Without diagnosis}& \textbf{Without prescriptions}& \textbf{Without procedures}& \textbf{Combined} \\ 
\hline
F1 score & 0.5233 ± 0.001& 0.5071 ± 0.002& 0.5275 ± 0.008& \textbf{0.5361 ± 0.003}\\ 
\hline
Accuracy & 0.7066 ± 0.000& 0.6964 ± 0.001& 0.7321 ± 0.004& \textbf{0.7321 ± 0.002}\\ 
\hline
Balanced accuracy & 0.7101 ± 0.001& 0.6958 ± 0.001& 0.7083 ± 0.006& \textbf{0.7166 ± 0.003}\\ 
\hline
Recall & \textbf{0.7165 ± 0.002}& 0.6947 ± 0.002& 0.6551 ± 0.010& 0.6885 ± 0.005\\ 
\hline
Precision & 0.4122 ± 0.000& 0.3993 ± 0.002& 0.4370 ± 0.006& \textbf{0.4389 ± 0.003}\\ 
\hline
AUROC & 0.7756 ± 0.001& 0.7699 ± 0.001& 0.7834 ± 0.000& \textbf{0.7930 ± 0.001}\\ 
\hline
AUPRC & 0.5058 ± 0.001& 0.4793 ± 0.002& 0.5162 ± 0.001& \textbf{0.5227 ± 0.002}\\ 
\hline
\end{tabular}%
}
\label{tab:ablation}
\end{table*}

\subsection{Benchmarking}
\begin{figure}[h]
    \centering
    \includegraphics[width=1\linewidth]{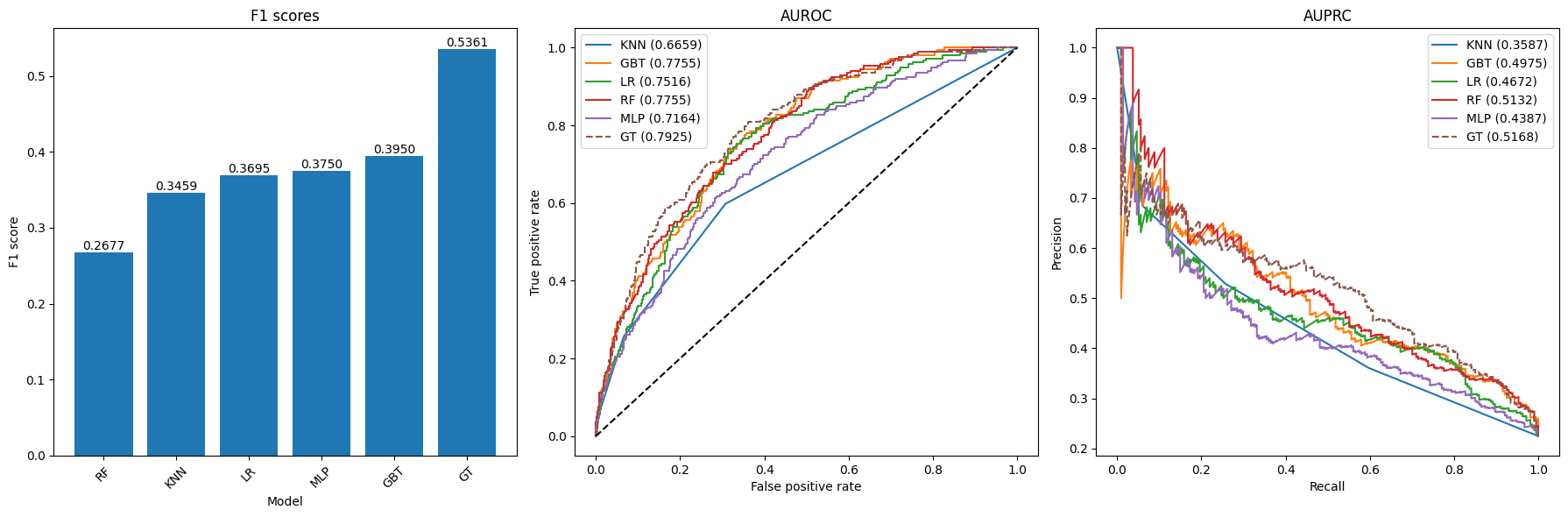}
    \caption{F1 scores, AUROC, and AUPRC curves of baseline algorithms on the test set, compared to the GT with FL, which utilizes relational information from the graph to make predictions. Standard deviations indicate variations across runs considering the same graph split.}
    \label{fig:metrics-baselines}
\end{figure}

We compared the performance of the GT with FL model against five baseline algorithms. The GT with FL model demonstrated an increased test AUROC (0.7925) and AUPRC (0.5168) compared to others (\autoref{tab:baselinemetrics-test}, \autoref{fig:metrics-baselines}). Although the differences in AUPRC between GT and Random Forest (AUPRC: 0.5132) were modest, the GT model's capacity to use graph-based relationships offers advantages. This is further investigated in the Discussion section.

\begin{table}[ht!]
    \centering
    \caption{Performance metrics (F1 score, AUROC, AUPRC) of baseline algorithms on the test set, compared to the GT.}
    \begin{tabular}{|c|c|c|c|}
        \hline
        \textbf{Algorithm} & \textbf{F1 Score} & \textbf{AUROC} & \textbf{AUPRC} \\
        \hline
        RF & 0.2677 & 0.7755 & 0.5132 \\
        \hline
        KNN & 0.3459 & 0.6659 & 0.3587 \\
        \hline
        LR & 0.3695 & 0.7516 & 0.4672 \\
        \hline
        MLP & 0.3750 & 0.7164 & 0.4387 \\
        \hline
        GBT & 0.3950 & 0.7755 & 0.4975 \\
        \hline
        GT & \textbf{0.5361} & \textbf{0.7925} & \textbf{0.5168} \\
        \hline
    \end{tabular}
    \label{tab:baselinemetrics-test}
\end{table}

\subsection{Interpretability Results}
\subsection{Graph descriptive statistics}
Our analysis focused on node degree and node similarity. TN and FP nodes exhibited the highest average degrees, indicating more diverse connections, while FN nodes had the fewest connections, suggesting that these HF patient profiles are more unique. Detailed metrics are available in the Supplementary Materials.

\subsection{Attention weights}
Attention weights, learned during training, highlight the importance of neighboring nodes' features for classifying a target node. We observed a bimodal distribution of weights in the final GT layer, indicating that the model assigned either high or low importance to neighbors.

Furthermore, TP and FP nodes, as well as TN and FN nodes, exhibited similar attention patterns. TN nodes assigned higher attention to other negative neighbors, helping with the correct classification, while TP nodes showed a more balanced attention across neighbor types. FN nodes resemble TN patterns but with slightly more attention to positive neighbors, indicating challenges in correct classification. Further details are available in the Supplementary Materials.

\begin{figure}[h]
    \centering
    \includegraphics[width=0.6\linewidth]{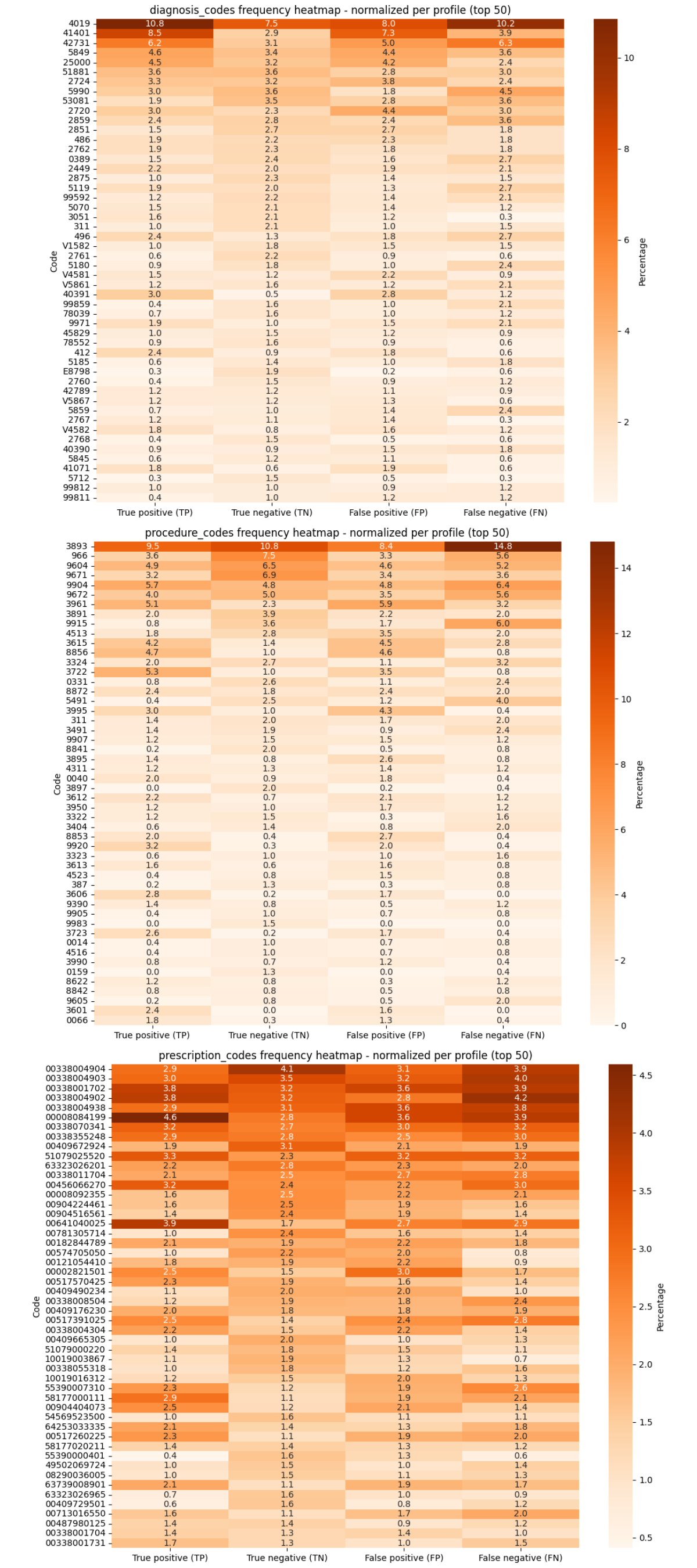}
    \caption{Heatmap of the top 50 most frequent codes across classification profiles from the test set, normalized by the number of patients in each profile. Medications show higher diversity across groups than other variables.}
    \label{fig:code-heatmaps}
\end{figure}

\subsection{Clinical features}
The clinical feature analysis refers to diagnosis, procedure, and prescription codes across classification profiles in the test set. The original patient data links to the embeddings used as node features in the similarity graph (see \autoref{fig:workflow}). \autoref{fig:code-heatmaps} shows a heatmap of the top 50 most frequent codes. The most important codes are translated in the main text and full code lookup is available in the Supplementary Materials, Table 12.

\textbf{Diagnoses:} The ICD-9 code for unspecified essential hypertension (4019) was the most prevalent across all profiles, especially in TP and FN, highlighting its association with HF. Atherosclerotic heart disease (41401) was also frequent in both TP and FP, indicating its significance as a HF marker but also its potential for leading to misclassifications. Atrial fibrillation (42731) was frequently observed in TP, FN, and FP profiles, indicating comorbidity. The higher frequency of chronic airway obstruction (496) and other respiratory-related diagnoses in FN suggests that patients with these complex comorbidities may be underdiagnosed for HF.

\textbf{Procedures:} Common critical care procedures such as endotracheal intubation (9604), mechanical ventilation (9672), and venous catheterization (3893) were commonly observed across all profiles. TP and FP profiles showed a higher occurrence of heart-related procedures, such as coronary artery bypass (3615) and coronary arteriography (8856), suggesting their importance in HF diagnosis but also their contribution to mislabeling. The FN profile had a unique pattern with increased frequencies of procedures like thoracentesis (3491) and parenteral infusion (9915), indicating that complex respiratory and nutritional management might lead to underdiagnosis.

\textbf{Prescriptions:} Prescription patterns were diverse, indicating the importance of pharmacotherapy in HF classification. Common hospital medications like intravenous sodium chloride (00338004904) and dextrose (00338001702) were frequently prescribed across profiles. In TP cases, high incidences of heparin sodium (00641040025) and potassium chloride (58177000111) were noted, emphasizing their role in managing HF. Furthermore, there was an overlap in medications between TP and FN profiles, including phenylephrine HCl (00517040525) and metoprolol (55390007310), suggesting that these patients receive similar pharmacological treatment indicative of their HF status.

\subsection{Integrative analysis}
\begin{figure}[h]
    \centering
    \includegraphics[width=0.95\textwidth]{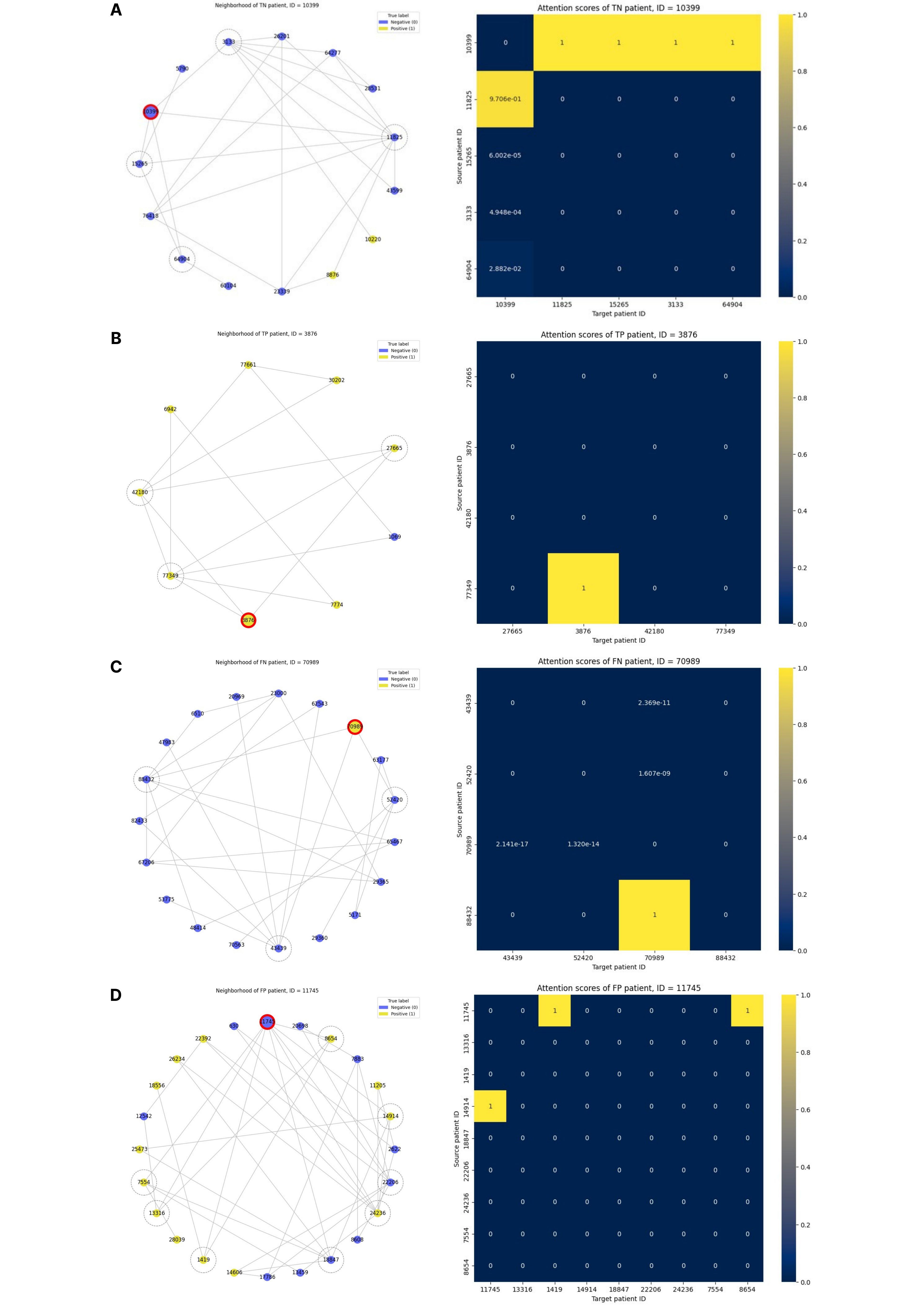}
    \caption{TN (A), TP (B), FN (C), and FP (D) instance information. Left: one and two hop neighborhoods of the central node, highlighted in red. Right: attention map with source and target nodes in the one-hop neighborhood of the central node.}
    \label{fig:integrative}
\end{figure}

A visualization of the immediate neighbors and the neighbors of neighbors (one and two-hop) of the four randomly selected nodes, along with the attention maps, is available in \autoref{fig:integrative}.

\textbf{True negative (TN):} The patient had strong similarities with neighboring negative patients, with uniformly high attention weights. The group shared non-cardiac conditions, such as persistent postoperative fistula, metabolic imbalances (e.g., acidosis), and chronic liver and kidney conditions. The shared procedures and medications, including long-term insulin therapy, exploratory laparotomy, and nystatin use, further reinforced the correct classification, as the patient showed a disease profile distinct from HF patients. Still, the model assigned a relatively high probability (0.4429) to the classification. This, combined with the presence of a few positive nodes within two hops, may indicate that the patient might have subtle HF risk factors.

\textbf{True positive (TP):} The patient had a profile resembling its positive neighboring nodes, all of whom were characterized by chronic cardiovascular diseases, diabetes (a common HF comorbidity), and ulcers. Shared features included advanced atherosclerosis, atrial fibrillation, and chronic kidney disease (another comorbidity) alongside procedures such as vascular bypass surgeries, arteriography, and toe amputation. Regarding medications, insulin and oxycodone were prescribed. The model assigned minimal attention to the neighbors, relying predominantly on the patient’s own features, likely due to the strong signals provided by their HF-related features.

\textbf{False negative (FN):} The patient had a neighborhood consisting entirely of TN patients who shared conditions such as severe infections and cancer. They had more unique diagnoses, including septicemia and breast cancer, diverging from the typical HF profile. Neighbor procedures like breast lesion excision and cancer-related surgeries also confirmed this divergence. Additionally, the patient’s medications were focused on antibiotics. Even with the low attention weights assigned to these neighbors, the model was unable to avoid misclassification. Thus, the patient's distinct profile probably caused the model to classify them as negative. This finding suggests a potential but less frequent connection between these conditions and HF.

\textbf{False positive (FP):}  The patient had a clinical profile similar to that of real HF patients. They shared cardiovascular conditions such as coronary atherosclerosis and coronary artery bypass surgery. That, combined with high attention weights assigned to positive neighbors, contributed to the misclassification. The patient’s diagnoses and procedures, which included essential hypertension, atrial fibrillation, and multiple cardiovascular surgeries, reinforced this profile. Misclassifications like this one could highlight individuals at a high risk of developing the disease. 

\section{Discussion}

 When comparing GNN architectures, the models demonstrated varying performances, with the Graph Transformer (GT) performing the best. This may be attributed to its advanced attention mechanism based on queries, keys, and values \cite{NIPS2017_3f5ee243}. Although GraphSAGE also showed a high AUPRC, this could be due to higher precision; however, precision is less critical than the higher recall shown by GT, as our scenario crucially requires identifying minority, HF-positive instances. Furthermore, all models benefited from loss functions adapted to class-imbalanced problems, such as focal loss and weighted cross-entropy, indicating they are indeed relevant for learning in unbalanced graphs.

Through the ablation study, prescription codes were found to be the most relevant class for correct predictions. This may be due to the higher proportion of medication data compared to procedures and diagnoses in the patient representations. In addition, these codes have been kept in the NDC standard, a choice that is often rare in the literature since these are often converted to more popular drug ontologies such as ATC. Our experiments suggest that using the full, raw NDC codes may offer advantages in terms of granularity for identifying different disease profiles.

In benchmarking, GT demonstrated superior F1 and AUROC scores. Nonetheless, similar AUPRC values for GT (0.5168) and RF (0.5132) suggest that RF, with appropriate threshold tuning, could also yield a higher F1 and serve as a resource-efficient alternative. However, GT’s graph-based approach provides advantages in uncovering subgroup-specific interactions, which RF does not capture, as it is restricted to the "bag-of-features" framework, limiting the obtention of relational insights.

Finally, the interpretability framework we introduced, based on graph descriptive statistics, attention weights, and clinical features, confirms the relevance of using graphs for healthcare tasks. By examining patient neighborhoods in the similarity graph, we were able to show that each classified patient requires careful analysis of its attributes, a process that may lead to uncovering clusters of high-risk patients or novel disease paths. Furthermore, this framework further reinforces that predictions in the healthcare field are inherently complex, far from simple (multi)classification scenarios. 

\section{Limitations}
Important limitations should be noted. First, the data resource was restricted to MIMIC-III, and data from other hospitals should also be evaluated in future studies. Furthermore, the single train-validate-test split may introduce selection bias, although experiments were repeated three times to help mitigate this. However, this limitation may be due to the emerging nature of the field of graph deep learning, as performing k-fold cross-validation in transductive settings is often not feasible due to library constraints.

Second, the use of ICD codes for HF labels introduces potential inaccuracies, as these codes may not fully capture the clinical nuances of HF. While we adhered to the official guidelines from the New York State's Department of Health, exploring other cohort identification methods could further improve label accuracy.

Moreover, as highlighted in prior studies \cite{jain2019attention}, the use of attention as an interpretability mechanism requires further investigation. Furthermore, inter-layer analyses should also be performed. 

Finally, reliance on a fixed threshold for calculating metrics like F1 is also a limitation. Future work could focus on optimizing AUROC for GNNs and baselines to enhance performance across varying thresholds.

\section{Conclusion}
The present study compared three GNN architectures (GraphSAGE, GAT, and GT) for predicting heart failure in an imbalanced patient similarity graph. The GT model, combined with focal loss, demonstrated the best performance. Through clinical feature ablation, medications were identified as the most relevant features. While GT's performance was comparable to Random Forests, its capacity to analyze relational data brought advantages for understanding the predictions. Furthermore, the graph interpretability analysis highlighted the importance of examining individual predictions, which may enable the identification of patients at high risk for developing heart failure and reveal novel patient profiles that may be less commonly associated with the disease.

Future work could investigate alternative graph representations, such as dynamic or heterogeneous graphs, as well as the potential of inductive graph learning to generalize predictions to new patients. Further enhancing the interpretability and explainability of GNNs with other axes is crucial for their integration into real-world clinical workflows. Moreover, optimizing decision thresholds could improve model performance and evaluation, particularly in scenarios with imbalanced datasets. Expanding the range of patient multimodal data, such as incorporating imaging and clinical notes while minimizing bias, will be essential to building more robust and reliable predictive models for healthcare applications.

\section*{Funding}
This work was financed in part by the Swedish Council for Higher Education through the Linnaeus-Palme Partnership, Sweden (3.3.1.34.16456), Coordenação de Aperfeiçoamento de Pessoal de Nível Superior (CAPES), Brazil - Finance Code 001, and Conselho Nacional de Desenvolvimento Científico e Tecnológico (CNPq), Brazil through grants nr. 309505/2020-8 and 308075/2021-8. We also acknowledge the support from Fundação de Amparo à Pesquisa do Estado do Rio Grande do Sul (FAPERGS), Brazil, through grants nr. 22/2551-0000390-7 (Project CIARS) and 21/2551-0002052-0.

\bibliographystyle{plain}
\bibliography{citation}

\end{document}